\documentclass[letterpaper,10pt]{extarticle}  

\usepackage{import}
\usepackage{local}
\usepackage{lipsum} 
\usepackage{times}  
\usepackage{helvet}  
\usepackage{courier}  
\usepackage{graphicx} 
\usepackage{amsmath} 
\usepackage{natbib}  
\usepackage{caption} 
\usepackage{algorithm}
\usepackage{algorithmic}
\usepackage{ragged2e}
\usepackage{newfloat}
\usepackage{listings}

\title{Fairness in Machine Learning Meets with Equity in Healthcare}

\author{%
\textbf{Shaina Raza},\textcolor{Accent}{\textsuperscript{1,*}} %
Parisa Osivand Pour,\textcolor{Accent}{\textsuperscript{1}} %
Syed Raza Bashir\textcolor{Accent}{\textsuperscript{2}}\\
\begin{small}
\textcolor{Accent}{\textsuperscript{1}} Vector Institute for Artificial Intelligence, Toronto, ON, Canada\\ 
\textcolor{Accent}{\textsuperscript{2}} Toronto Metropolitan University, Toronto, ON, Canada\\
\textcolor{Accent}{\textsuperscript{*}}Correspondence: \textcolor{Accent}{shaina.raza@vectorinstitute.ai} \\
\end{small}
}

\begin{document}

\begin{justify}
\maketitle

\begin{abstract}
With the growing utilization of machine learning in healthcare, there is increasing potential to enhance healthcare outcomes. However, this also brings the risk of perpetuating biases in data and model design that can harm certain demographic groups based on factors such as age, gender, and race. This study proposes an artificial intelligence framework, grounded in software engineering principles, for identifying and mitigating biases in data and models while ensuring fairness in healthcare settings. A case study is presented to demonstrate how systematic biases in data can lead to amplified biases in model predictions, and machine learning methods are suggested to prevent such biases. Future research aims to test and validate the proposed ML framework in real-world clinical settings to evaluate its impact on promoting health equity.
\end{abstract}

\section{Introduction}
Machine learning (ML) offers immense potential to significantly enhance patient outcomes and transform the landscape of clinical healthcare \cite{thomasian2021advancing}. 
Utilizing its analytical and predictive capabilities, ML can help reveal disease patterns and trends, and optimize patient care. However, it is important to proceed with caution when leveraging ML in healthcare. This is because inherent biases and inequalities in the data may result in discrimination, which could result in worsening of pre-existing health disparities \cite{mcneely2020social}. For example, a model trained on biased data might inaccurately predict a higher risk of heart disease for specific racial or ethnic groups, leading to unequal treatment opportunities and poorer health outcomes \cite{obermeyer2019dissecting}. 

In the context of  ML, the term “bias” refers to skewed outcomes caused by errors in the modeling process \cite{fletcher2021addressing} . This often occurs when training data is unrepresentative or contains systemic errors, leading the model to learn and potentially replicate these biases in its predictions.  “Disparity” in healthcare indicates inequalities in health status, healthcare access, or healthcare quality across different groups \cite{raza2022machine}. It is very important to minimize these biases and disparities when applying ML to healthcare, to ensure equitable outcomes.

Health equity \cite{sikstrom2022conceptualising} is a core principle in clinical healthcare that seeks to eliminate differences in health outcomes and access to equal healthcare among various populations. This principle aims to ensure that all individuals, regardless of their demographic or socio-economic background, have equal opportunities to access care and maintain or improve their health. Both the World Health Organization (WHO)  and the United Nations (UN) prioritize health equity as a critical element of their missions to enhance global health outcomes. This motivates us to pursue research in this domain.

In this study, we introduce an Artificial Intelligence (AI) framework designed to ensure that ML models produce unbiased and equitable predictions for all populations. Specifically, we integrate software engineering principles into the framework to improve its modularity, maintainability, and scalability, making it adaptable and efficient for various applications. Our goal is to introduce fairness in the healthcare setting through ML. The term fairness typically refers to the algorithm’s ability to make decisions and predictions without unjust bias or discrimination \cite{rajkomar2018ensuring}. Fairness is a critical aspect of responsible AI and ML practices, especially in sensitive areas like healthcare. 

We put forth a fair ML framework, in this work, that is rooted in software engineering principles. Following that, we present a healthcare case study that demonstrates how biases can exacerbate disparities in healthcare access and outcomes. We also detail how our suggested framework can aid in advancing health equity. It is our hope that integrating this framework into healthcare systems will promote equal health opportunities and outcomes.

\section{Previous Works}
In the realm of healthcare, researchers have explored the potential of AI and its capacity for ensuring fairness and equity. Rajkomar et al. \cite{rajkomar2018ensuring} highlighted the importance of fairness in clinical care and introduced research guidelines and technical solutions to combat bias-es through ML . Fletcher et al. \cite{fletcher2021addressing} conducted research on the global health context, particularly in Low- and Middle-Income Countries (LMICs), proposing three criteria—appropriateness, fairness, and bias—to evaluate ML for healthcare. Raza \cite{raza2022machine} presented a review on the challenges for ML within a general view of public health and its influences. Thomasian et al. \cite{thomasian2021advancing} urged for policy-level consensus on algorithmic bias and providing principles for mitigating bias in healthcare. Wesson et al. \cite{wesson2022risks} looked the potential benefits and drawbacks of using big data in research, emphasizing the importance of an equity lens in health. 

Sikstrom et al. \cite{sikstrom2022conceptualising} conducted literature survey on fairness in AI and ML, striving to operationalize fairness in medicine.  Concurrently, Gervasi et al. \cite{gervasi2022potential} explored fairness, equity, and bias in ML algorithms within the health insurance industry. Obermeyer et al. \cite{obermeyer2019dissecting} uncovered racial bias in a commercial algorithm used for identifying high-risk patients, emphasizing the need to address racial bias in ML pipelines.  The AI Now Institute \cite{ainow} delved into the social implications of AI and ML, publishing works on fairness, accountability, and transparency, including healthcare ML pipelines. Google AI for Social Good program \cite{googleai_socialgood} are also developing tools and resources like the What-If Tool and Fairness Indicators to assist practitioners in identifying and mitigating biases in ML pipelines \cite{Huang2021, raza2022dbias, sikstrom2022conceptualising}. These works high-light the growth of ML in healthcare. Nevertheless, there is need for engagement with fairness, bias, and ML processes in healthcare.

Our work differentiates from previous research by offering a practical, end-to-end framework for implementing fair ML in healthcare. Our work is rooted in software engineering principles and focuses on specific fairness considerations in healthcare. Different from much of the existing work that mainly focuses on theoretical guidelines or principles, our research offers a tool that can have real-world applicability and continuous improvement.
\section{Proposed Framework}
We propose an AI framework, shown in Figure 1 and the steps given in Algorithm 1, that integrates software engineering principles with fairness in ML. The goal of this work is to enhance modularity, maintainability, and scalability. The steps of our proposed framework are as: 

\begin{figure*}
    \centering
    \includegraphics[width=0.75\linewidth]{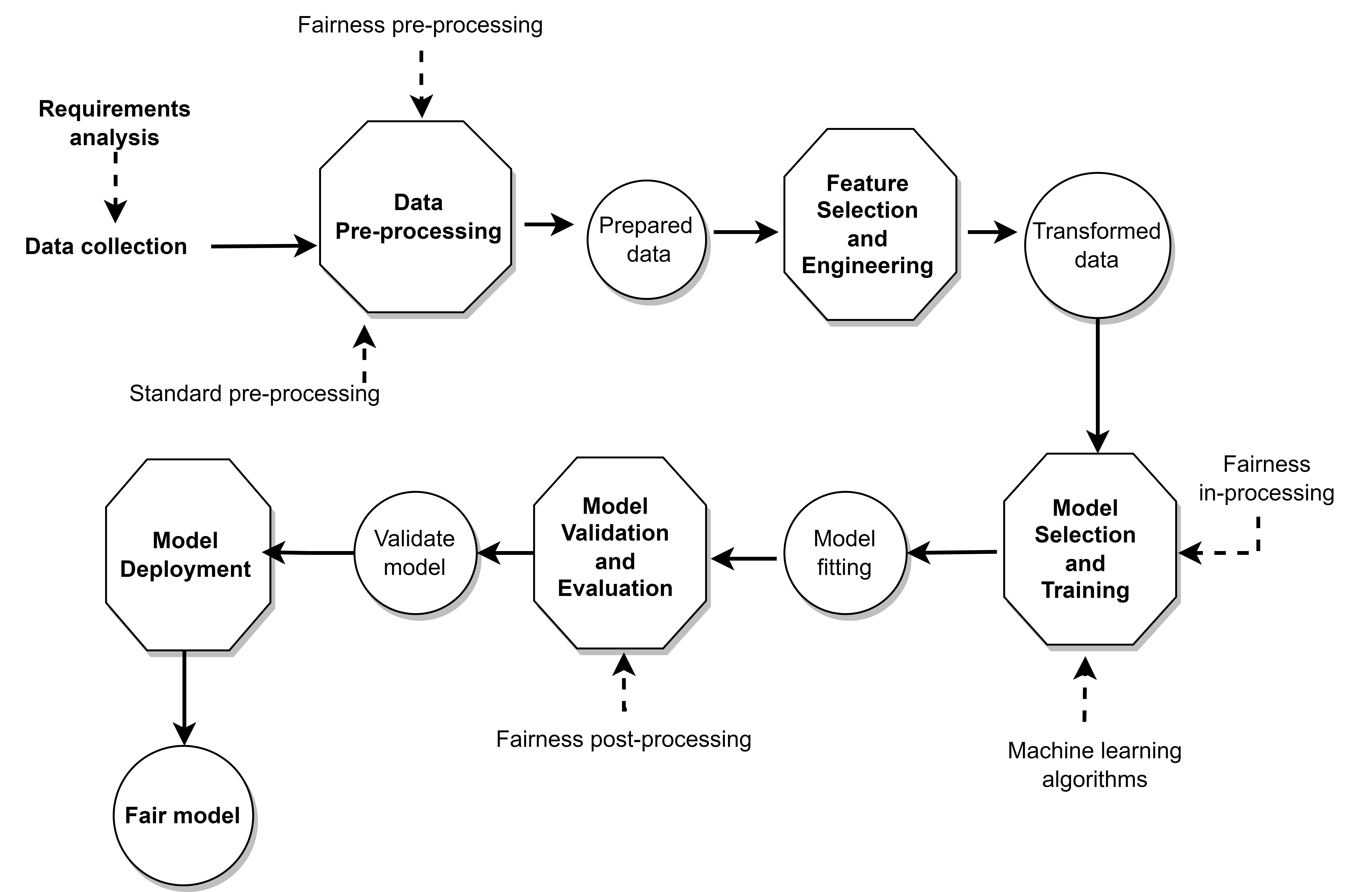}
    \caption{Proposed Framework}
    \label{fig:fig1}
\end{figure*}

\begin{algorithm}[tb]
\caption{Fairness in ML for healthcare}
\label{alg:fairness_healthcare}
\textbf{Input}: Data set D, ML algorithm A\\
\textbf{Parameter}: Fairness metric F, Threshold T\\
\textbf{Output}: Trained ML model M, Fairness evaluation E
\begin{algorithmic}[1]
\STATE Pre-process data set D to get $D'$.
\STATE Split $D'$ into training, validation, and test sets: $T_{\text{train}}$, $T_{\text{val}}$, and $T_{\text{test}}$.
\STATE Apply fairness pre-processing to get $T_{\text{train\_balanced}}$.
\STATE Initialize ML algorithm A with parameters P.
\STATE Train ML model M using $T_{\text{train\_balanced}}$.
\STATE Apply fairness in-processing on M during training.
\STATE Validate M on $T_{\text{val}}$, evaluate performance metrics and fairness metric F.
\WHILE{$F > T$}
\STATE Tune hyperparameters P of ML algorithm A.
\STATE Retrain M using $T_{\text{train\_balanced}}$ and updated P.
\STATE Validate M on $T_{\text{val}}$, re-evaluate performance metrics and fairness metric F.
\ENDWHILE
\STATE Evaluate M on $T_{\text{test}}$ to get final performance metrics and fairness metric F.
\STATE Apply fairness post-processing on M if necessary.
\STATE Deploy M in production environment.
\STATE Monitor M performance and fairness metric F on new data.
\IF {$F > T$}
\STATE Update M and repeat steps 8-13.
\ENDIF
\STATE \textbf{return} M, E
\end{algorithmic}
\end{algorithm}

\textit{Actor Identification}: Identify the key actors and their roles is one of the fundamental steps in software engineering principles. Understanding the users is crucial in this process as it provides context and direction for the subsequent steps in the framework.

\textit{Requirements Analysis}: Identify the problem that we aim to solve in healthcare and determine the fairness requirements specific to the context. For example, to understand the ethical, legal, and social implications of the solution \cite{cordeiro2021digital, lu2022considerations} and set goals to mitigate potential biases and promote equitable outcomes.

\textit{Data Collection:} Collect diverse and representative data samples that covers various demographic \cite{tramer2017fairtest}, to ensure the model generalizability. It is also important to ensure data privacy and security standards are met while acquiring and storing data \cite{raza2023constructing,v14122761}

\textit\textit{Data Pre-processing:} Apply best practices to clean, normalize, and transform the data. Implement fairness pre-processing techniques such as re-sampling \cite{drummond2003c4}, re-weighting \cite{Kamiran2012}, or editing feature values \cite{hardt2016equality} to reduce potential biases in the data set.

\textit{Feature Selection and Engineering: }Identify relevant features that impact the target outcome and avoid features that might introduce biases \cite{ahmad2018interpretable}. Apply domain knowledge to create meaningful features that contribute to a fair model.

\textit{Model Selection and Training:} Choose a suitable ML algorithm for the problem at hand, considering software engineering principles like modularity, scalability, and maintainability \cite{raza2022dbias}. Employ in-processing techniques such as fair classification \cite{Kamiran2012}, clustering \cite{chierichetti2017fair}, adversarial learning \cite{madras2018learning}, and counterfactual fair learning \cite{kusner2017counterfactual} algorithms to promote fairness during model training.

\textit{Model Validation and Evaluation: }Assess the model performance using standard evaluation metrics as well as fairness-specific metrics like disparate impact \cite{feldman2015certifying}, demographic parity \cite{madras2018learning}, or equalized odds \cite{golz2019paradoxes}. Optimize the model using hyperparameter tuning, and apply post-processing fairness techniques like counterfactual analysis \cite{kusner2017counterfactual} or calibration \cite{NIPS2017_b8b9c74a} to adjust the predictions as needed.

\textit{Model Deployment and Monitoring: }Deploy the model in a production environment while adhering to software engineering best practices for continuous integration, continuous deployment, and monitoring \cite{alanazi2022using}. Regularly evaluating the model performance on new data, ensuring its fairness and generalizability over time is also important. Given that software engineering relies on empirical science for validity, understanding the users and their feedback is also important to validate our framework approach.

\section{Case Study}
\textit{Case study:} We take a case study \cite{ting2017development} of Diabetic retinopathy (DR) using ML. DR is a complication of diabetes that can lead to vision loss and blindness if not detected and treated early. The prevalence of diabetes continues to rise globally, and early detection of DR is crucial for timely intervention and preventing vision loss. Fundus photography is a widely used technique to screen for DR, but the manual examination of these images can be time-consuming and subject to inter-observer variability. ML algorithms can automate this process, making it more efficient and consistent, as shown in this work \cite{ting2017development}.

\textit{Objective:} Our objective in this paper is to propose a fair and unbiased version of the original ML method \cite{ting2017development} for early detection of diabetic retinopathy.

\textit{Data Collection and Pre-processing: }A diverse and representative data set of fundus images was collected from various sources, ensuring the inclusion of different demo-graphic groups such as age, gender, and ethnicity. The data was pre-processed, including cleaning, normalization, and transformation. Fairness-enhancing pre-processing techniques: re-sampling \cite{Huang2021} and re-weighting \cite{feldman2015certifying} were applied to balance the data set and mitigate potential biases.

\textit{Feature Selection and Engineering: }Domain experts identified relevant features for the detection of diabetic retinopathy, such as blood vessel structure, hemorrhages, and microaneurysms. Feature engineering techniques were applied to extract meaningful information from fundus images while avoiding features that might introduce bias.

\textit{Model Selection and Training:} A convolutional neural network (CNN) \cite{mallat2016understanding} was selected as the ML algorithm, considering its effectiveness in image analysis tasks and alignment with software engineering principles like modularity and scalability. Fair classification \cite{dwork2012fairness} and adversarial learning \cite{zhang2018mitigating} techniques were applied during the model training to ensure fairness and unbiased predictions.

\textit{Model Validation and Evaluation:} The model was evaluated using standard metrics such as accuracy, precision, and recall, as well as fairness-specific metrics like demographic parity and equalized odds. Post-processing fairness techniques like counterfactual analysis \cite{kusner2017counterfactual} and calibration \cite{NIPS2017_b8b9c74a} were applied as needed to adjust the predictions and ensure fairness across demographic groups.

\textit{Model Deployment and Monitoring: }The CNN model was deployed in a production environment for continuous integration, continuous deployment, and monitoring. The model's performance and fairness were regularly evaluated on new data, and updates were made as needed to maintain its fairness and generalizability over time.

\textit{Outcome}: The fair ML-based system improved the efficiency and consistency of DR screening, reducing the workload of healthcare professionals and enabling timely intervention. By ensuring equitable predictions across diverse demographic groups \cite{tramer2017fairtest}, the system contributed to health equity and reduced the risk of vision loss in diabetic patients \cite{raza2022machine}.
\section{Discussion}
The proposed framework and its application to the DR case study illustrate the promising potential of integrating software engineering principles with fairness in ML for healthcare. In this context, we highlight several critical observations and implications.

\textit{Achieving Fairness in Machine Learning:} The integration of fairness-enhancing techniques during pre-processing, in-processing, and post-processing stages were essential to mitigate potential biases and ensure fairness in the DR detection model. These techniques worked in harmony with standard ML processes, ensuring their applicability in other healthcare contexts. However, it's worth noting that fairness is a dynamic concept, dependent on the specific healthcare problem and demographics \cite{dwork2012fairness}. Therefore, the selection and application of fairness techniques require domain knowledge and an understanding of the specific fairness requirements.

\textit{Role of Software Engineering Principles:} Adopting software engineering principles not only enhanced the modularity, maintainability, and scalability of our ML framework but also facilitated the integration of fairness techniques. By treating the ML model as a software product, we were able to design and develop the framework more efficiently and effectively. This allowed us to monitor and adjust the model continuously, ensuring that fairness and performance were maintained over time.

\textit{Health Equity in Practice:} By developing a fair and unbiased DR detection model, we demonstrated the practical application of health equity in ML for healthcare. The model made equitable predictions across diverse demographic groups, contributing to equal health opportunities and outcomes for diabetic patients. This underscores the importance of fairness in healthcare ML, not only as a theoretical concept but also as a practical tool for promoting health equity.

\textit{Collaboration Across Disciplines:} This work exemplified the benefits of collaboration across disciplines. The integration of insights and methodologies from software engineering, ML, healthcare, and ethics was essential to the successful development of the proposed framework. This underscores the importance of interdisciplinary collaboration in addressing complex problems like fairness in ML for healthcare.

Despite the encouraging results from our framework and the DR case study, several challenges and limitations need to be addressed in future work. For example, it's essential to understand the legal and ethical implications of deploying such frameworks, especially regarding data privacy and informed consent \cite{tramer2017fairtest}. Additionally, the selection and definition of fairness metrics can be subjective and may differ across contexts, requiring further investigation. 

In future work, we envision refining and expanding our proposed framework to address these challenges. We also plan to apply our framework to other healthcare problems to further validate its efficacy and versatility. Furthermore, we hope to stimulate discussions and collaborations with stakeholders from various disciplines to contribute to the ongoing research and development of fair ML for healthcare. Ultimately, our goal is to contribute to health equity and improve patient outcomes using ML, while upholding the principles of fairness and justice.

\section{Conclusion}
We present an approach aimed at uncovering and addressing biases present in healthcare data, with the goal of promoting equitable solutions. The case study examined provides a foundation, suggesting that the proposed framework can effectively identify biases and apply suitable fairness methods to assess potential discrimination and generate fairer outcomes. To maximize benefits from this framework, it is crucial to prioritize fairness in all aspects of model design, deployment, and evaluation. This study has some limitations, for example, the lack of real-world empirical evidence supporting the effectiveness of the proposed AI framework. Further empirical research and real-world validation are needed to verify the proposed framework efficacy.

\section{Acknowledgments}
Resources used in preparing this research were provided, in part, by the Province of Ontario, the Government of Canada through CIFAR, and companies sponsoring the Vector Institute \url{www.vectorinstitute.ai/#partners}

\bibliographystyle{plain}
\bibliography{localbibliography}
\end{justify}
\end{document}